\documentclass[conference]{IEEEtran}

\usepackage{graphicx}
\usepackage{amsmath}
\usepackage{subcaption}
\usepackage{siunitx}

\ifCLASSINFOpdf
\else
\fi
\usepackage{amsmath}
\begin{document}

\title{TVIM: Thermo-Active Variable Impedance Module: Evaluating Shear-Mode Capabilities of \textit{Polycaprolactone} }

\author{Trevor~Exley,~\IEEEmembership{Member,~IEEE,}Rashmi~Wijesundara,~\IEEEmembership{Member,~IEEE,}Shuopu~Wang, Arian~Moridani, and~Amir~Jafari,~\IEEEmembership{Member,~IEEE}
\thanks{T. Exley and A. Jafari are with Advanced Robotic Manipulators (ARM) Lab, the Department of Biomedical Engineering, University of North Texas, Texas, United States}\\
{$^\ast$To whom correspondence should be addressed; E-mail:  amir.jafari@unt.edu..}
\thanks{Manuscript received April 19, 2005; revised August 26, 2015.}}

%
%

\markboth{Journal of \LaTeX\ Class Files,~Vol.~14, No.~8, August~2015}%
{Shell \MakeLowercase{\textit{et al.}}: Bare Demo of IEEEtran.cls for IEEE Journals}
%



\maketitle

\begin{abstract}
In this work, we introduce an advanced thermo-active variable impedance module which builds upon our previous innovation in thermal-based impedance adjustment for actuation systems. Our initial design harnessed the temperature-responsive, viscoelastic properties of Polycaprolactone (PCL) to modulate stiffness and damping, facilitated by integrated flexible Peltier elements. While effective, the reliance on compressing and the inherent stress relaxation characteristics of PCL led to suboptimal response times in impedance adjustments.
Addressing these limitations, the current iteration of our module pivots to a novel 'shear-mode' operation. By conducting comprehensive shear rheology analyses on PCL, we have identified a configuration that eliminates the viscoelastic delay, offering a faster response with improved heat transfer efficiency.
A key advantage of our module lies in its scalability and elimination of additional mechanical actuators for impedance adjustment. The compactness and efficiency of thermal actuation through Peltier elements allow for significant downsizing, making these thermal, variable impedance modules exceptionally well-suited for applications where space constraints and actuator weight are critical considerations.  
This development represents a significant leap forward in the design of variable impedance actuators, offering a more versatile, responsive, and compact solution for a wide range of robotic and biomechanical applications.

\end{abstract}

\begin{IEEEkeywords}
Variable Impedance Module, VIA, polycaprolactone, thermoresponsive polymer, Peltier.
\end{IEEEkeywords}

%
\IEEEpeerreviewmaketitle

\section{Introduction}

%
%
%
%


\IEEEPARstart{\textit{P}}{\textit{hysical}} human-robot interactions (\textit{p}HRI) are becoming more and more common as robotic platforms aim to be more prevalent in society in the form of assistive devices, wearable technologies, and prosthetics\cite{haddadin_safe_2011,pervez_safe_2008,duchaine_safe_2009}. Throughout these interactions, applications often deal with a variety of unknown and changing environments. Traditional actuation has been deemed insufficient for \textit{p}HRI due to high reflected inertia and lack of backdrivability \cite{haddadin_physical_2016}. Safer operations and interactions require robotic systems to include various levels of impedance. Impedance refers to the resistance of motion with three main components: stiffness, damping, and inertia. Variable impedance actuators (VIA) enable safer robotic operations and interactions by dynamic modification of the impedance by adjusting the impedance parameters, including changing stiffness, damping ratios, and even inertia to suit evolving conditions. 

To better improve \textit{p}HRI, researchers have investigated different designs of actuation technology, specifically for tasks involving contact or impact. For more than 25 years, series elastic actuators (SEA) have been successfully implemented in a wide variety of applications\cite{hurst_series_2020,herr2007artificial,parietti_series_2011,marquardt_design_2021}. This actuation technology provides many benefits in force controls of robots with advantages such as increased peak power output, passive energy storage, and low output impedance. Despite this, SEAs only offer control of impedance at the design stage meaning the fixed compliance determined by the mechanical properties of the elastic element. This greatly reduces the adaptability of SEAs in certain situations where adjustments to impedance parameters are required. 

Work have been done across the domain of VIA with various subsets implemented across all fields, one such is the variable stiffness actuator (VSA). However, VSAs are limited in the adjustment of only stiffness of the output link during physical interaction with external environments. Moreover, variable damper actuators and variable inertia actuators have significantly contributed to the field\cite{laffranchi2010variable,baser_employing_2020,laffranchi_compact_2011}. However, they have garnered less attention in robotics research and \textit{p}HRI due to the absence of energy-saving and release components, which are present in VSAs\cite{jafari_awas-ii_2011}. Consequently, this has led to a diminished potential for reducing energy consumption in periodic motions. However, damping proves to be essential in applications where maintaining the energy of the system within specific thresholds is essential for stability, as seen in components such as the knee and ankle joints of bipedal robots\cite{pieringer_review_2017,milazzo_modeling_2024}.

\begin{figure}[t]
    \centering
    \includegraphics[width=.9\columnwidth]{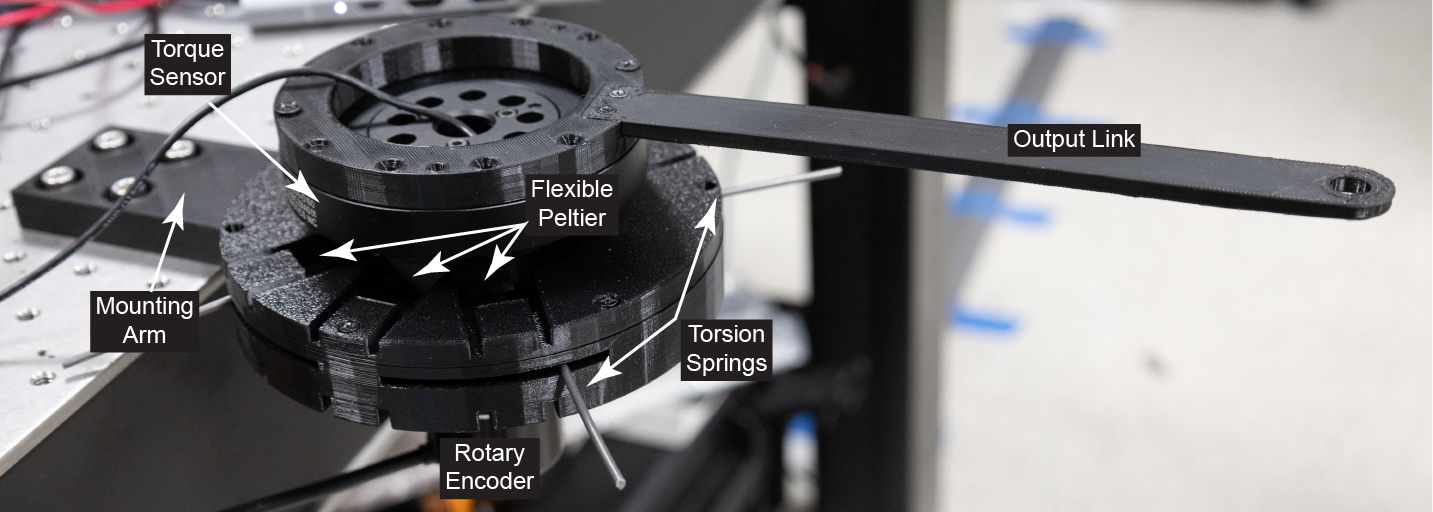}
    \caption{Shear-mode thermo-active VIA prototype constructed with 3D-printed PETG components with integrated Peltiers.}
    \label{fig:vimreal}
\end{figure}

Each VIA should be able to control at least two degrees of freedom (i.e. link's trajectory and impedance) and required with two independent sources of mechanical power, generated from sources such as electric motors. Additionally, a third motor will be required in the case that both damping and stiffness needs to be controlled, independently, in addition to the link trajectory. However, the implementation of these mechanisms makes the overall actuator component to be bulky, heavy, and large in comparison to SEAs or traditional actuators. Thus, these factors limit their suitability and adaptability in many applications. Though there are strategies to decrease the size of VIAs, these approaches would reduce their applications and adaptability in many situations. 

Previously, we have introduced a novel thermal-based VIA with a unique adjustment in the actuator's impedance by controlling the temperature of a thermoplastic polymer, \textit{polycaprolactone} (PCL), instead of a dedicated mechanical mechanism. PCL is a thermoplastic polymer that exhibit changes in visco-elasticity as they transition from a cold (rigid) to a hot (soft) state. To control the temperature, flexible Peltiers are embedded into our previous design to adjust the impedance, in a compact and lightweight realization. However, a major limitation of such design is that compressing a relatively large body of PCL leads to an increase in backlash and further slows impedance adjustment due to the time-dependent characteristics of stress relaxation and creep. By redesigning the interaction between the output link and PCL, these detrimental effects can be minimized in the context of impedance regulation.

This paper presents a new design which works based on shear forces between the frame and viscoelastic polymer housed in a Variable Impedance Module. The organization of this article is as follows: mechanical design, stiffness formulation and rheology modeling are presented in Section II, while section III discusses the results on rheology measurements and perturbation testing. Section IV presents discussion and future works.


\section{Mechanical Design and Rheology}

\subsection{Mechanical Design}

The earlier VIA design \cite{exley_novel_2023,exley_toward_2023} encountered problems related to backlash and stress-relaxation characteristics, resulting in dis-connection between the variable impedance module and output link. To address these issues, a shear-mode design is proposed. This design aims to enhance heating efficiency and eliminate undesirable effects.

\begin{figure}
    \centering
    \begin{subfigure}[b]{0.475\textwidth}
        \centering
        \includegraphics[width=.95\textwidth]{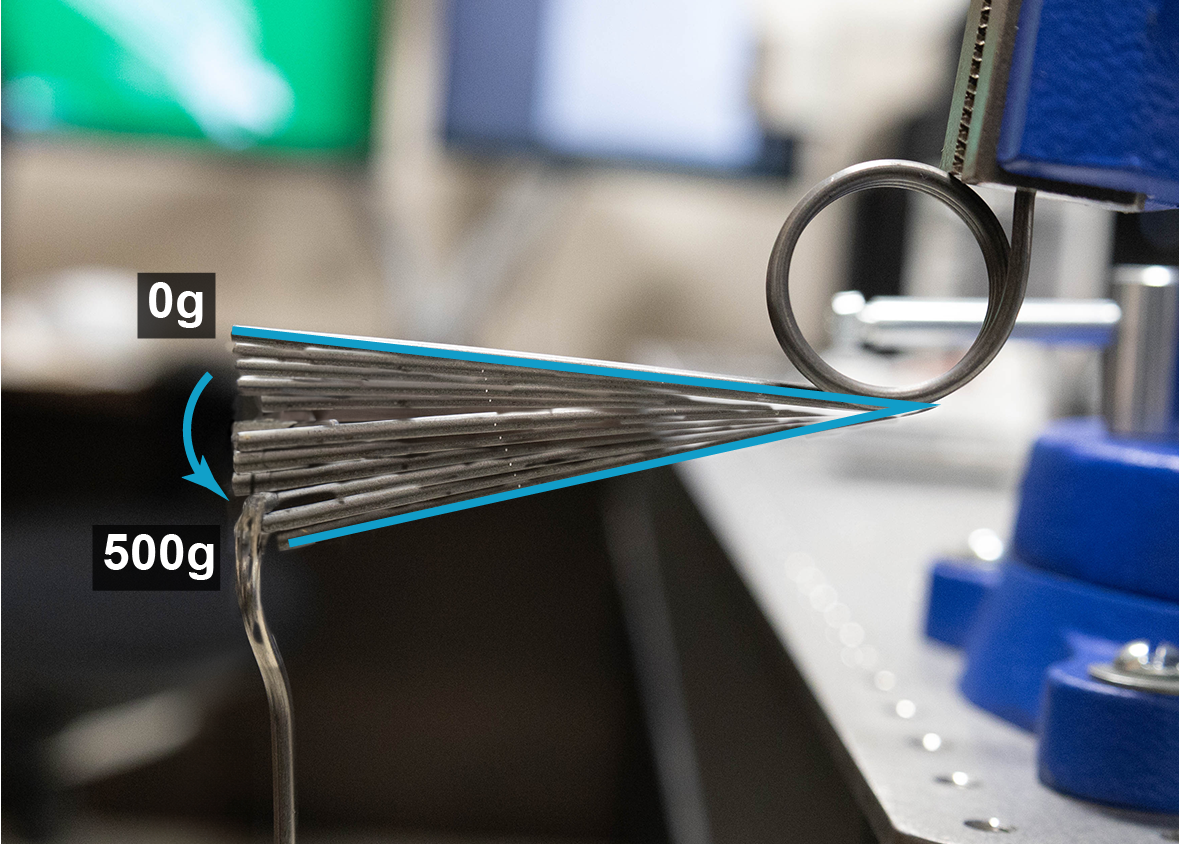}
        \caption{Multiple exposure of torsion spring deflection with added hanging weight}
        \label{fig: torsionspring}
    \end{subfigure}
    \hfill
    \begin{subfigure}[b]{0.475\textwidth}
        \centering
        \includegraphics[width=.95\textwidth]{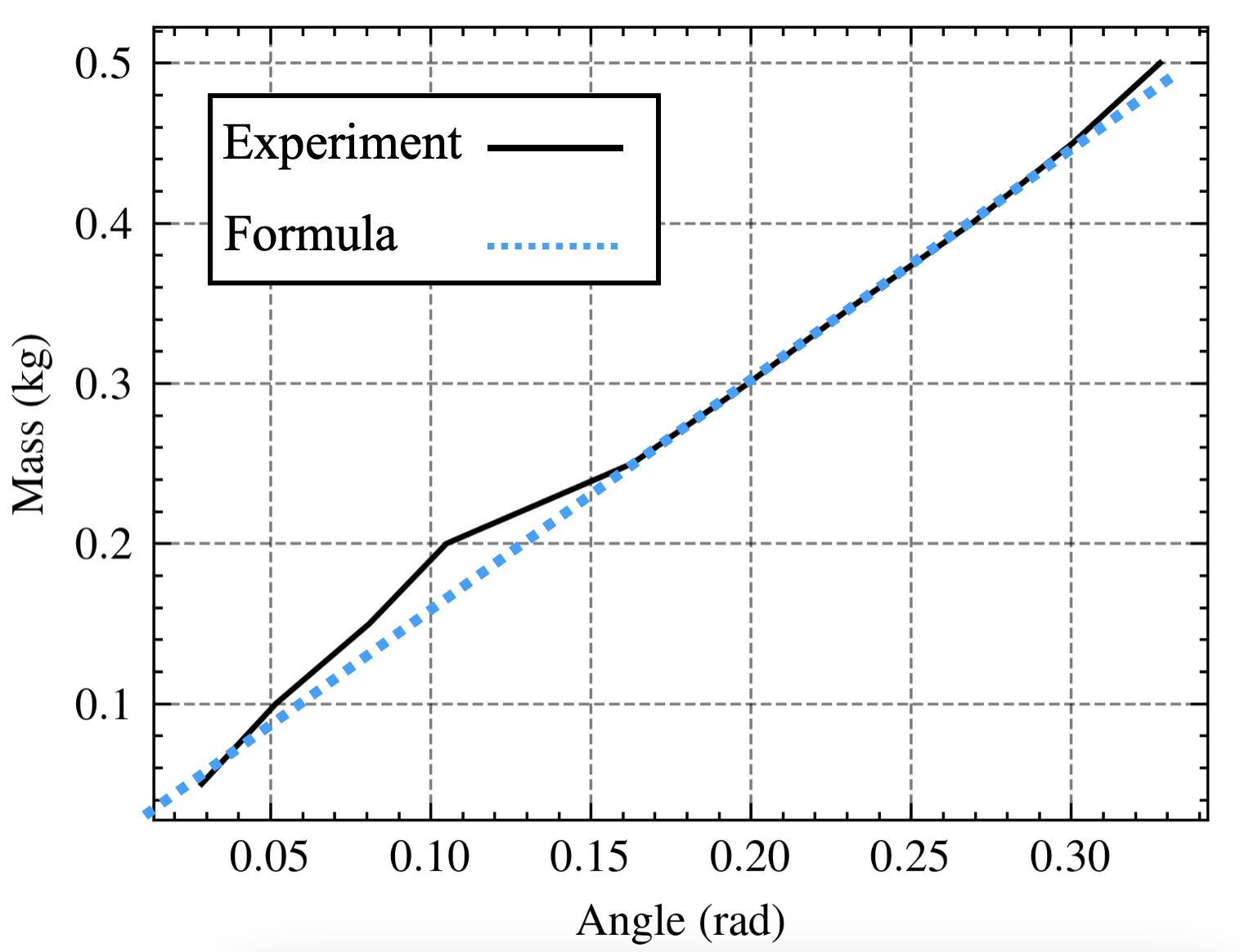}
	    \caption{Graph of mass vs angle for torsion spring}
	    \label{fig: torsionmass}
	    \vspace{0.0cm}
    \end{subfigure}
    \caption{Determination of torsion spring stiffness using hanging weights- the slop represents the stiffness $k_{torsion}$}
    \label{fig:torsionstiff}
\end{figure}

\begin{figure}[t] 
\centering 


\begin{subfigure}[h]{0.4\textwidth} 
    \includegraphics[width=.95\textwidth]{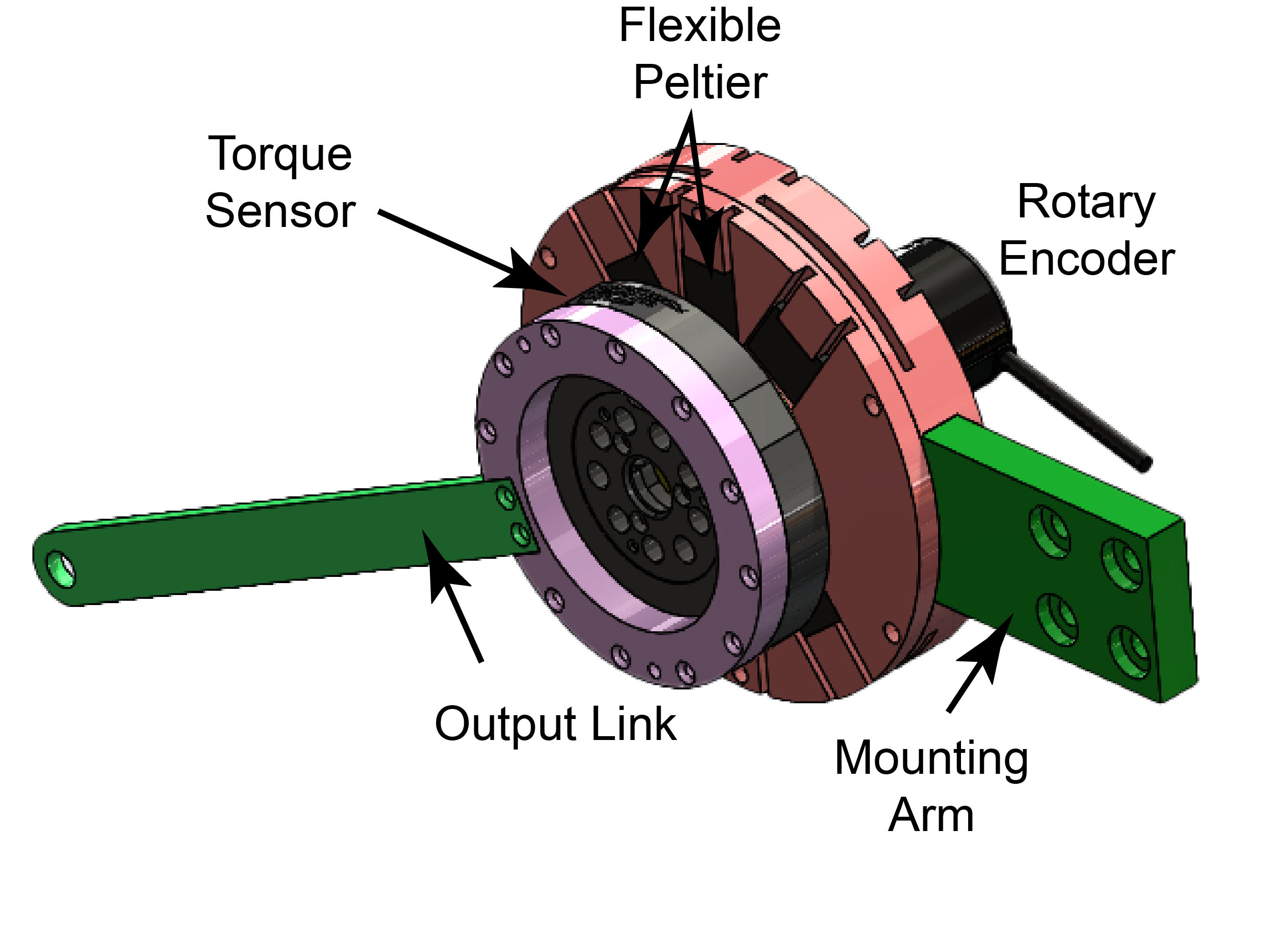} 
    \caption{Isometric view}
    \label{fig:labCADiso}
\end{subfigure}
\hfill 


\begin{subfigure}[h]{0.35\textwidth}
    \includegraphics[width=.95\textwidth]{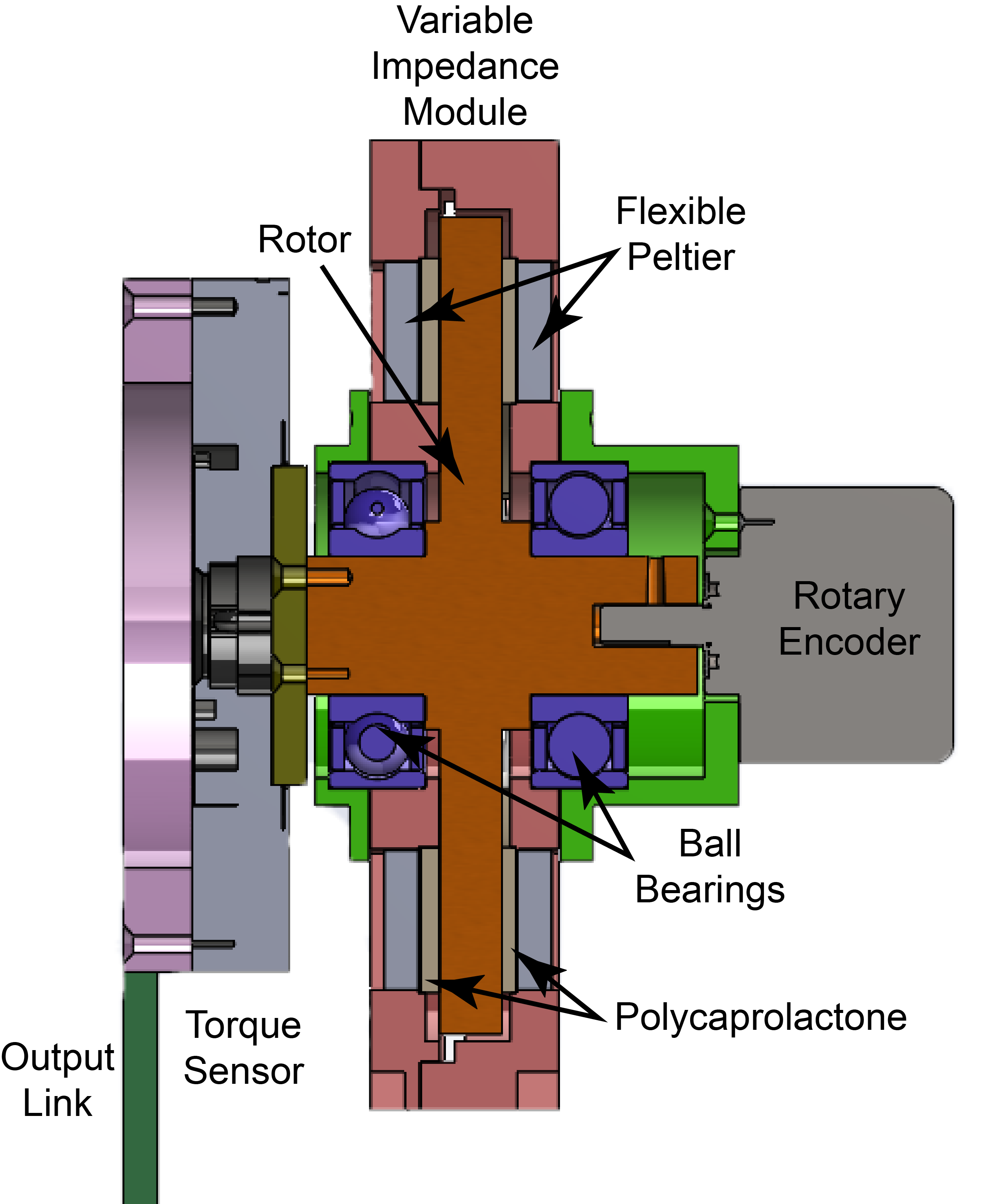}
    \caption{Side view}
    \label{fig:labCADpcl}
\end{subfigure}
\hfill 

\begin{subfigure}[h]{0.35\textwidth}
    \includegraphics[width=.95\textwidth]{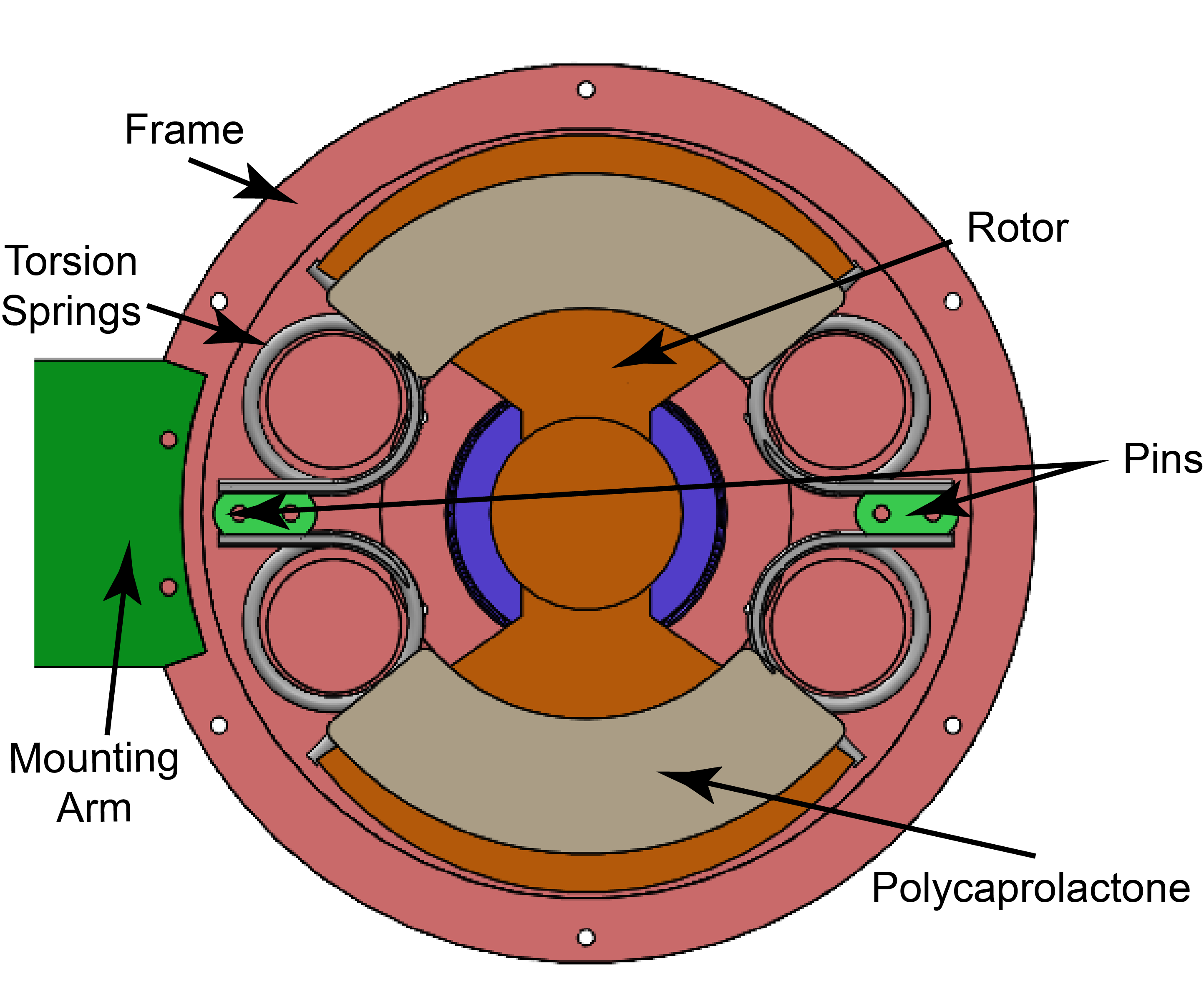}
    \caption{Internal view}
    \label{fig:labCADside}
\end{subfigure}

\caption{Labeled CAD views of the Thermo-active VIM} 
\label{fig:labCADviews} 
\end{figure}

In the shear-mode thermo-active variable impedance module (Fig. \ref{fig:vimreal}), the stiffness of torsion springs is a critical aspect for its mechanical performance. The module integrates a cylindrical core and planar rotor that is in contact with four torsion springs, as shown in Figure \ref{fig:labCADside}. These springs connect the stationary housing to an output link, ensuring constant stiffness through interactions with the rotor. The calculation of torsion spring stiffness $k_{torsion}$ involves the spring's physical and material properties, such as the Young's modulus ($E$), wire diameter ($d$), mean coil diameter ($D$), and the number of active coils ($n$). The stiffness is derived using the following equation:

\begin{equation}
k_{torsion} =\frac{E \times d^4}{64 \times D \times n}
\end{equation}

This formula allows for the precise calculation of the spring's resistance to torsional deformations, essential for the module's intended function. This derivation was confirmed measuring deflection while applying a known torque (Fig. \ref{fig:torsionstiff}).

\begin{table}[ht]
    \centering
    
    \begin{tabular}{|c|c|} \hline 
    \textbf{Mechanical Parameter} & \textbf{Value} \\ \hline
    Young's Modulus, $E$ & 210 GPa \\ \hline
    Wire diameter, $d$ & 2.413 mm \\ \hline
    Mean coil diameter, $D$ & 24.4094 mm \\ \hline
    Number of coils, $n$ & 3.25 \\ \hline
 Spring stiffness, $k_{torsion}$& 12.5 Nm/rad\\\hline
 Moment of Inertia&415.6 $kg/mm^2$\\\hline
 Distance of springs from center axis & 57 mm \\ \hline
 Maximum angular deflection &$\pm0.104$ rad\\ \hline
 PCL contact area & 1579.2 mm$^2$ \\ \hline
    \end{tabular}
    \caption{Mechanical parameters of torsion spring and shear-mode VIM}
    \label{tab:spring_parameters}
\end{table}

The overall stiffness of the output link due to the springs, $K_s$ is a function of torsional springs' stiffness $k_{torsion}$ and other geometrical factors such as the distance between center of springs and rotor center or rotation. 


In our design, there exists a direct correlation between the angular rotation of the rotor (denoted as $\alpha$) and the angular deflection of the spring (denoted as $\theta$) for rotation angles smaller than 0.2 radians
From the torsional spring constant, we can derive the torsional torque as:
\begin{equation}
    T_{\text{torsion}} = k_{\text{torsion}} (\theta - \theta_{\text{pretension}})
    \label{eq:torsion_torque}
\end{equation}
where \( \theta \) is the angle of twist, and \( \theta_{\text{pretension}} \) is the pretension angle in the spring.

This equation is for an individual torsion spring and assumes that the force applied on the rotor can be represented by a compression spring.


Here, \( K_{\text{s}} \) represents the effective stiffness of the output link, which is influenced by the torsional stiffness of the spring and the radius at which the force acts.

The total energy stored in the springs during the deflection of the output link (i.e., rotation of the rotor) should equate to the potential energy of the output link attributed to the springs. Given that the stiffness of the torsion spring was verified as a constant value in Figure \ref{fig:torsionstiff}, and similarly, the output link stiffness was confirmed in Figure \ref{fig:perturb}, it is reasonable to employ the Energy method as described in Equation \ref{eq:energy} to ascertain the output link stiffness as a function of $\theta$, particularly for small rotor rotations of less than 0.2 radians, as:

\begin{equation} \label{eq:energy}
    E = 4(\frac{1}{2}k_{torsion}\alpha^2) = \frac{1}{2}K_s\theta^2
\end{equation}
This equation relates the spring stiffness $k_{\text{torsion}}$ to the output link stiffness $K_s$, utilizing both the spring deflection $\alpha$ and the projection angle to the rotor $\theta$. As previously noted, there exists a one-to-one correspondence between $\alpha$ and $\theta$ (i.e., $\frac{\alpha}{\theta}=1$). This relationship can be manipulated to derive the output link stiffness as a function of the spring stiffness:

\begin{equation} \label{eq:energyderived}
    K_s(\theta) = 4k_{torsion}{\theta^2}
\end{equation}

The equation of motion for the output link as a result of the output torque is as follows:

\begin{equation}
T_L = b_{\text{PCL}}(\dot{\theta}) - (K_s + K_{PCL})(\theta) + I_l\ddot{\theta}
\end{equation}
where $I_l$ is the link's inertia, $b_{PCL}$ is the damping of the PCL, $K_{PCL}$ is the stiffness of the PCL, and $K_s$ is the total spring stiffness.

All calculated and derived mechanical parameters can be found in Table \ref{tab:spring_parameters}.

\begin{figure}[h]
    \centering
    \includegraphics[width=.9\columnwidth]{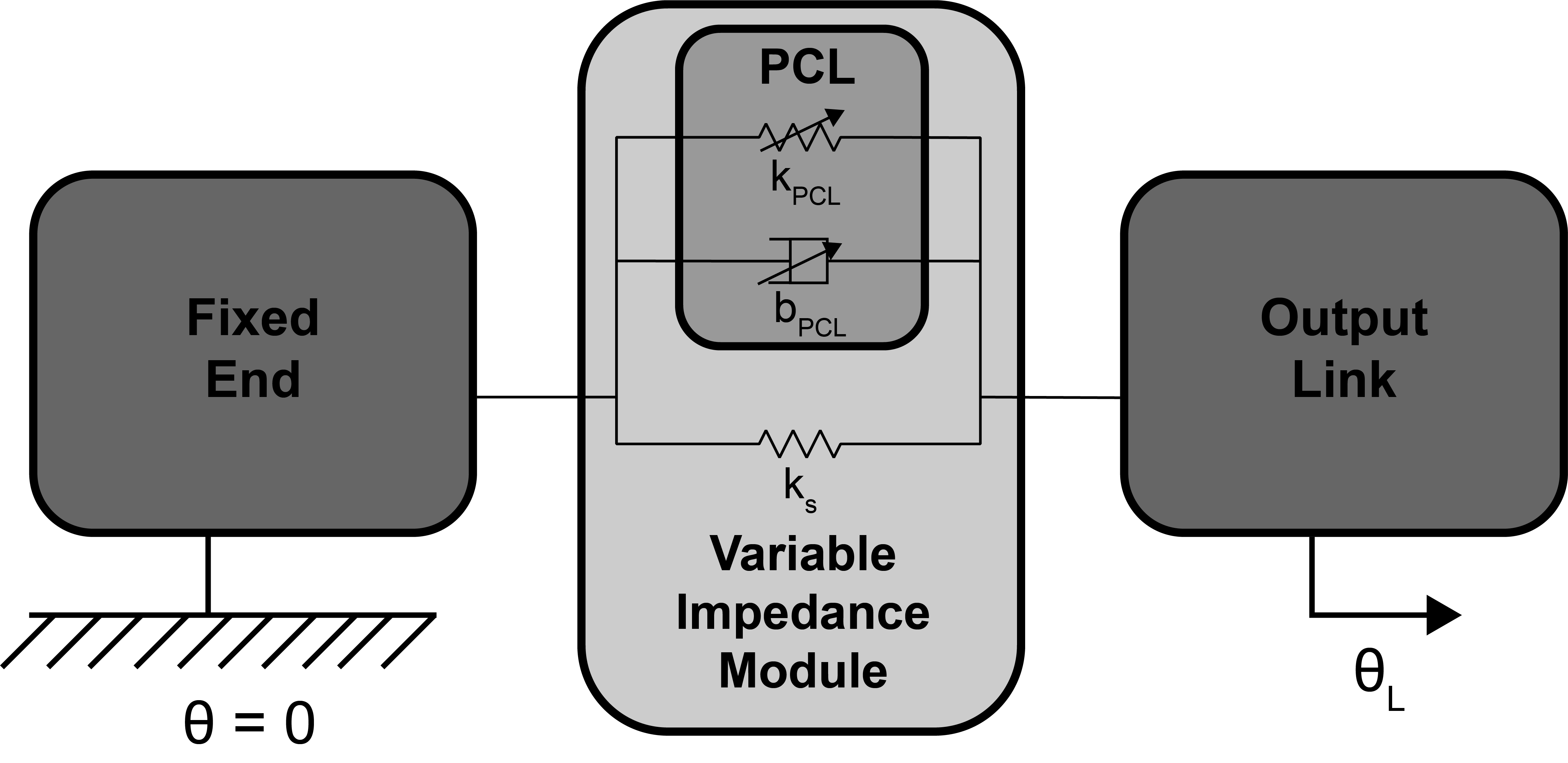}
    \caption{Fixed end mechanical model of the VIM, showing constant stiffness from torsion springs, and variable stiffness and damping from the PCL.}
    \label{fig:enter-label}
\end{figure}

\begin{figure}[h] 
\centering 

\begin{subfigure}[b]{0.45\textwidth} 
    \includegraphics[width=0.9\columnwidth]{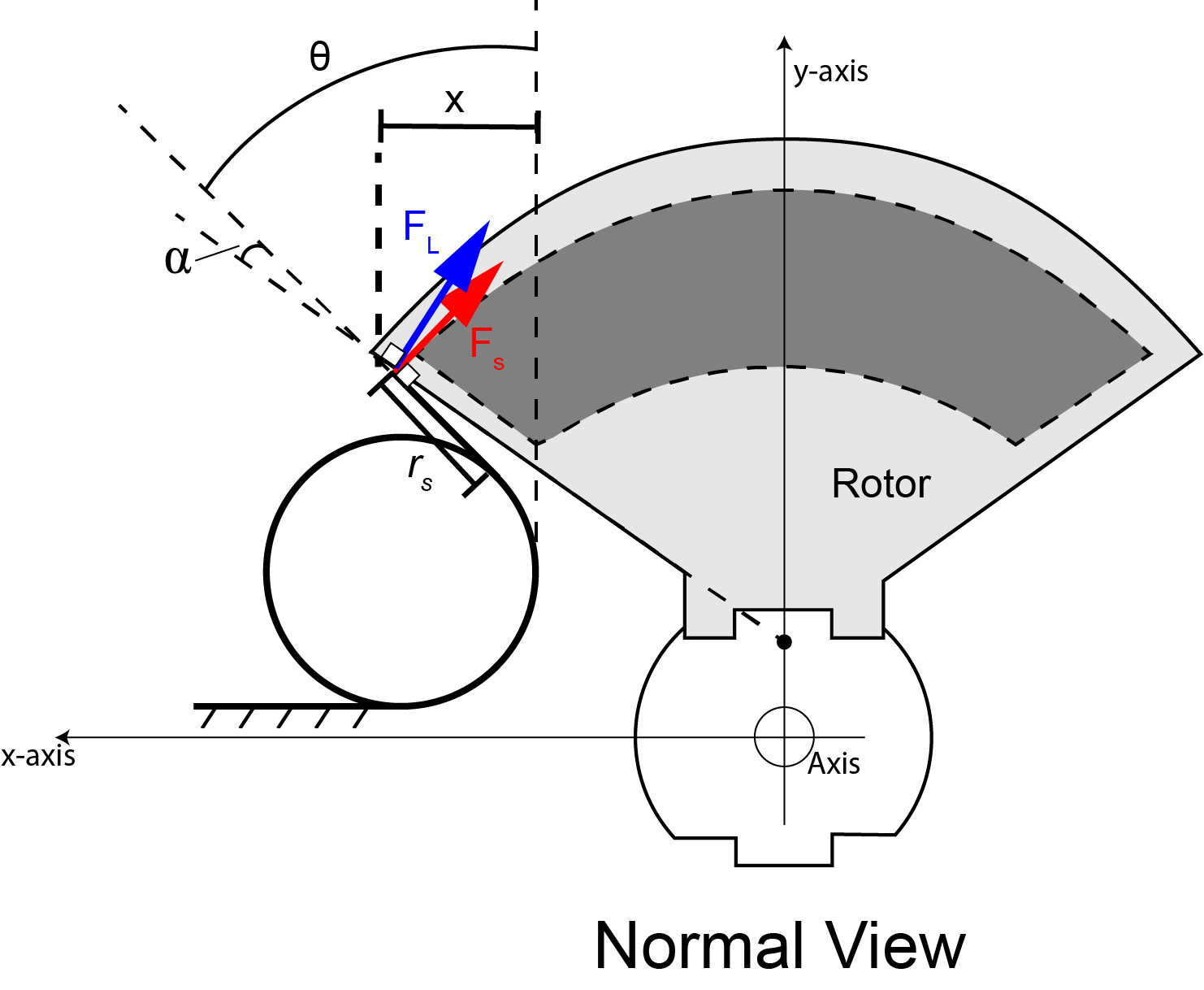} 
    \caption{Normal View}
    \label{fig:model1}
\end{subfigure}
\hfill 

\begin{subfigure}[b]{0.45\textwidth} 
    \includegraphics[width=0.9\columnwidth]{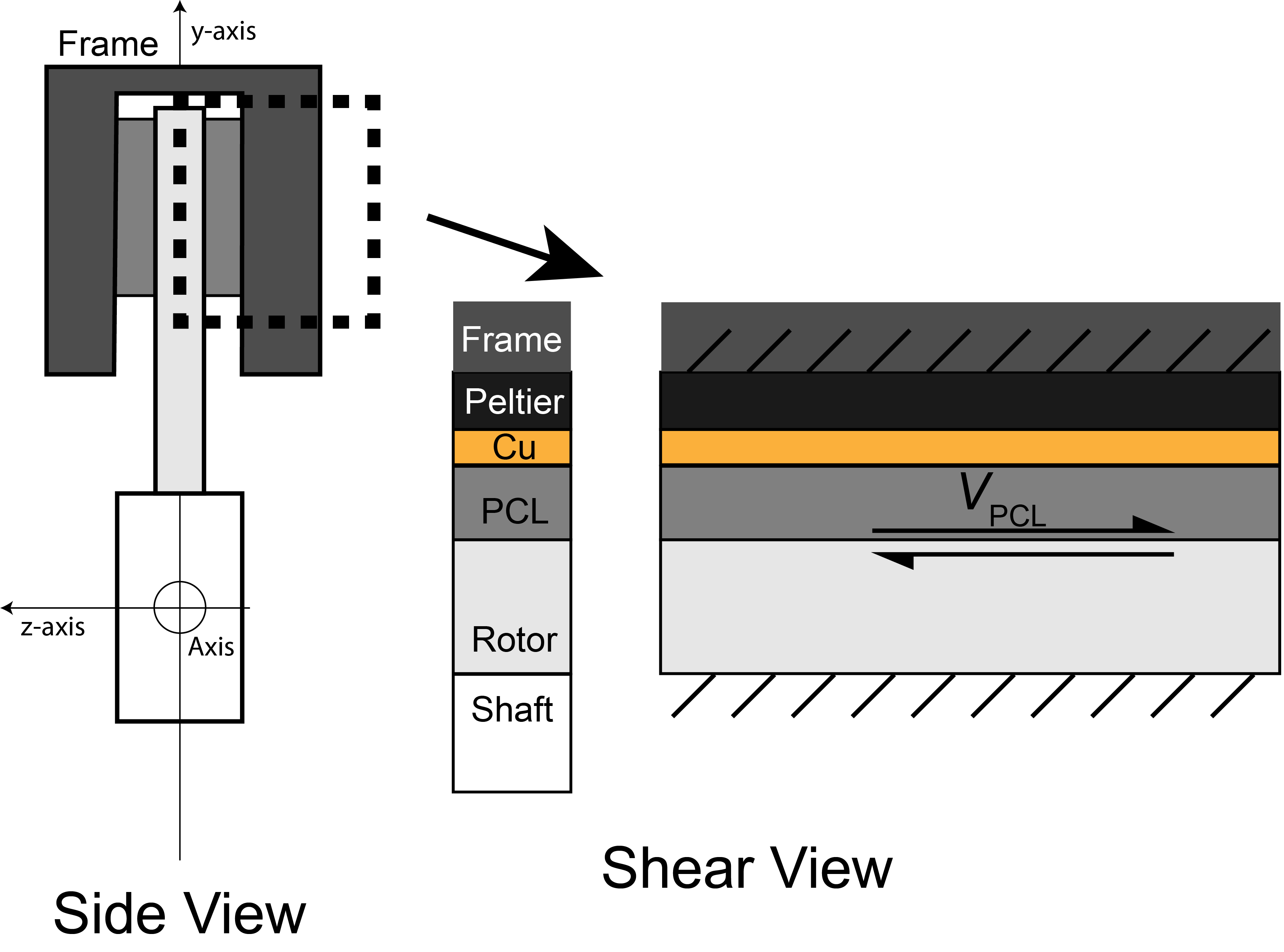} 
    \caption{Side and Shear view}
    \label{fig:model2}
\end{subfigure}

\caption{Mechanical model showing (a) contact forces between torsion spring as rotor turns projected to the central axis, and (b) shear forces between the PCL and the rotor. } 
\label{fig:models} 
\end{figure}

\subsection{Rheology}

\begin{figure*}[t]
    \centering
    \includegraphics[width=.9\textwidth]{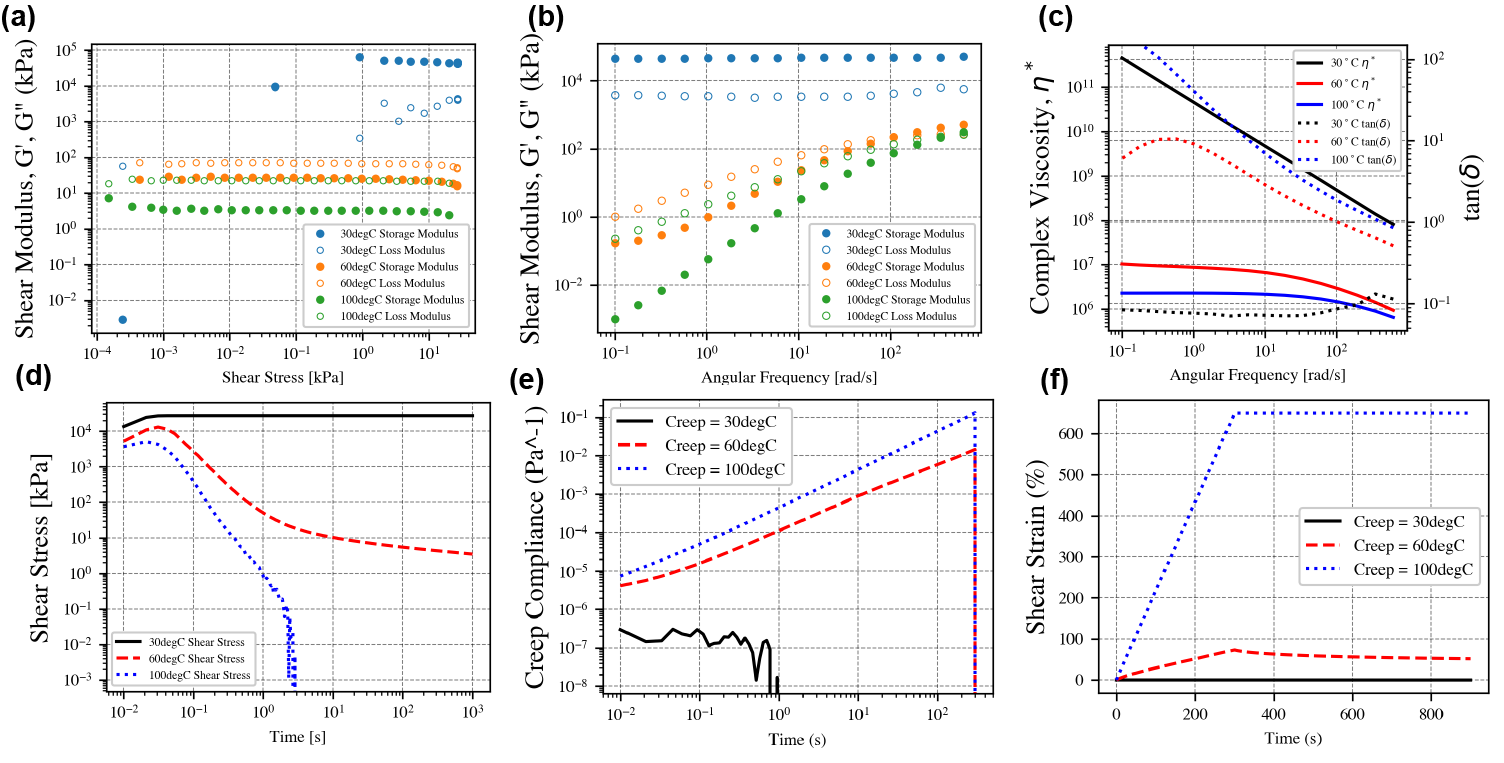}
    \caption{Rheology data of PCL across temperatures of 30$^\circ$C, 60$^\circ$C, and 100$^\circ$C. a)Amplitude sweep, b,c) frequency sweep, d,e) creep, and f) stress-relaxation.}
    \label{fig:modfreq}
\end{figure*}

Polycaprolactone, often used for its bio-compatibility and versatility in biomedical applications, lacks rheological behavior analysis in actuator integrations. 

Rheological tests (MCR-92; Anton Paar, Austria) covered creep, stress relaxation, frequency sweep, amplitude sweep, and temperature sweep (-5$^\circ$C to 200$^\circ$C). The set of tests was conducted examining PCL behavior at different temperatures (30$^\circ$C (below), 60$^\circ$C (at), and 100$^\circ$C (above) melting temperatures). In each trial, heating commenced at 60$^\circ$C until complete melting, with the PP25 parallel plate (D = 25 mm) adjusted to a 1 mm clearance.


\begin{figure}
    \centering
    \includegraphics[width=.95\columnwidth]{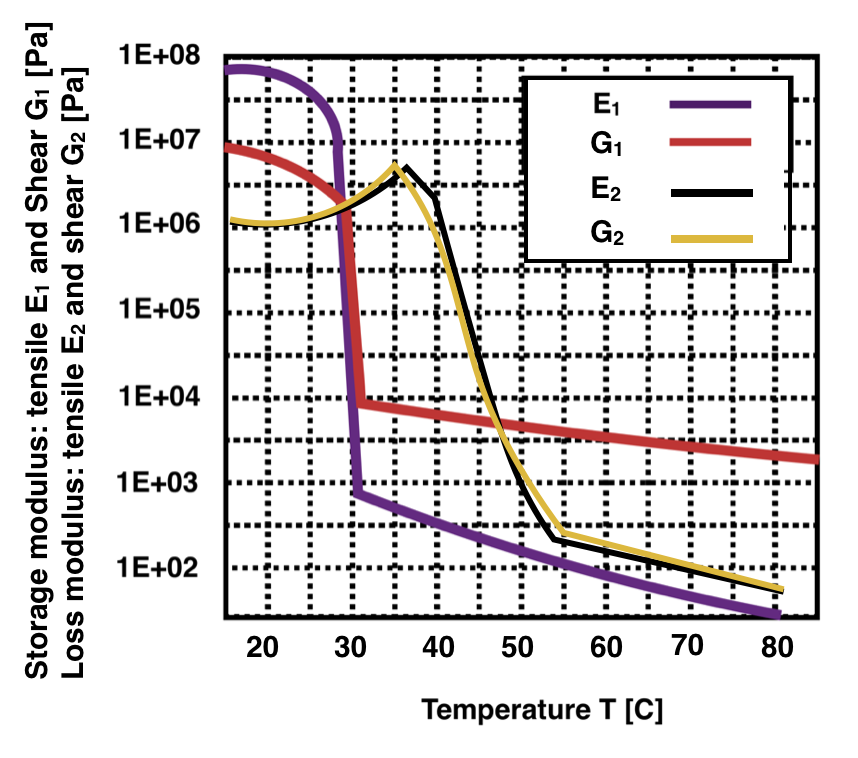}
    \caption{Storage and Loss (tensile and shear) Moduli of PCL at different temperatures (Adapted from Mosleh et al.)\cite{mosleh_tpupclnanomagnetite_2013}}
    \label{fig:tempresponse}
\end{figure}
 Rheological evaluation of PCL began with amplitude sweep tests to identify linear viscoelastic region followed by frequency sweep tests (0.1 $\frac{\text{rad}}{\text{s}}$ to 628 $\frac{\text{rad}}{\text{s}}$) using a 0.01 \% oscillation strain (Fig. \ref{fig:modfreq}a). Subsequently, creep and stress relaxation tests (constant shear strain rate of 10 \%) further probed the samples' viscoelastic properties through application of a 50 Pa shear stress for 300 seconds followed by a 200 second relaxation period to record the resulting shear stress.



\begin{equation}
    \|\eta^*\| = \frac{\sqrt{G'^2 + G"^2}}{\omega}
\end{equation}

The complex viscosity $\|\eta^*\|$ [Pa$\cdot$s] \cite{soares_chapter_2015} (Fig. \ref{fig:modfreq}c) measures how much a material resists flow under an applied force and its ability to deform under force.
The loss factor $tan(\delta)$ indicates the damping property of a material, representing the ratio of its viscous to elastic response of a material under oscillatory deformation, providing insights on a material's methods of dissipating energy. These two parameters show two main findings: (i) as temperature increases, the material becomes more substantially less viscous, demonstrating enhanced damping, and (ii) at higher temperatures as frequency increases the loss moduli decrease to magnitude of 10$^{-1}$ indicating more elastic behavior.
The creep compliance \( J(t) \) [Pa\(^{-1}\)] (Fig. \ref{fig:modfreq}d-e) characterizes a material's time-dependent deformation under constant load, vital for predicting actuator behavior \cite{wilczynski_brittle_2021}. In the creep test, a 50 Pa shear stress was applied for 300 s, followed by a 600 s of relaxation. These profiles offer crucial insights into PCL's deformation behavior under sustained mechanical loads, essential for VIA applications. For example, as a sustained load is applied to the PCL at 30$^\circ$C, the material acts as a solid, while at increased temperatures there is a gradual increase in the deformation, leading to further damped systems.




\section{Results}

\subsection{System Identification}

\begin{equation}
    (k_{PCL} + k_s)\theta_l = I_l\ddot\theta_l +b_{PCL}\dot\theta_l+ T_l
\end{equation}

In our proposed design, the fixed springs around a rotor can be modeled as springs in parallel to PCL which can be modeled as a combination of a spring and a dashpot. There are two main methods to model a viscoelastic material such as PCL, namely; Maxwell \cite{johnson1992viscohyperelastic}, or Kelvin-Voigt \cite{hosford2010mechanical} models. Both models use springs to depict a pure elastic element and a dashpot to represent a pure viscous element.  

In the Maxwell model, both the spring and the dashpot are subject to the same stress, but each element has an independent strain. In the Kelvin-Voigt model, both the spring and the dashpot are subjected to the same strain but each element has independent stress. The Maxwell model can predict the recovery phase (after releasing the external load) more precisely than the Kelvin-Voigt model. However, the Kelvin-Voigt model can more accurately represent the strain rate-time dependency which is fundamental in viscoelastic behaviour, i.e. the creep phenomenon \cite{vizesi2008stress}. In modeling a viscoelastic material with pre-dominant elasticity, the Kelvin-Voigt model is usually applied to describe the creep behavior of the material due to its practicality and wide application in the engineering field\cite{rivlin1992elasticity}.

In our design, we present the PCL as a Kelvin-Voigt model, since the compression springs will be modeled as a spring in parallel with PCL, which will have higher stiffness than that of the spring in the PCL model. 

\begin{equation} \label{eq:kelvinvoigt} \sigma (t)= E_1 \epsilon (t)+\eta(G_2) \frac{d\epsilon(t)}{dt},
\end{equation}

\begin{figure}[h]
    \centering
    \includegraphics[width=.9\columnwidth]{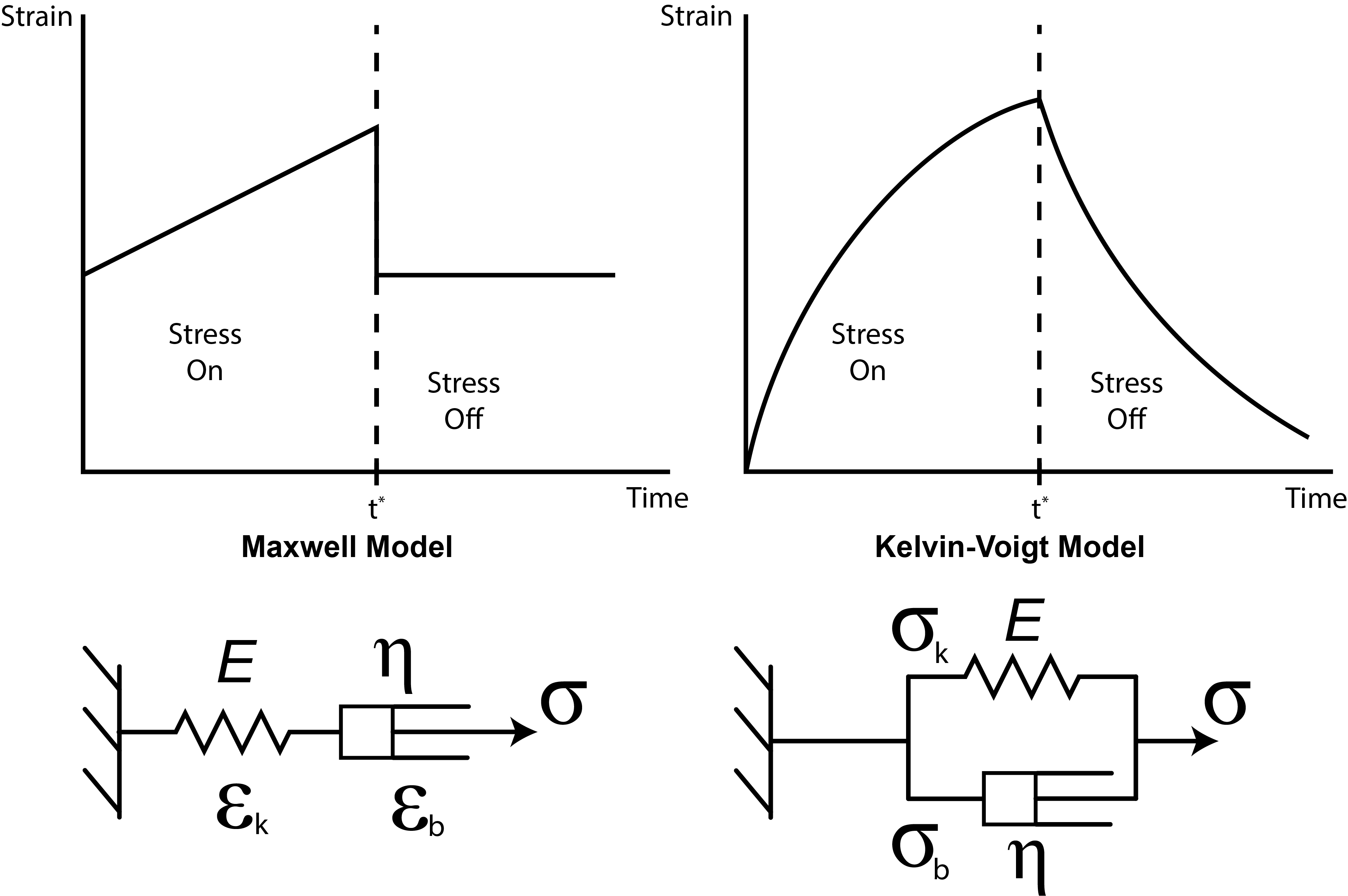}
    \caption{Maxwell vs Kelvin-Voigt models showing stress relaxation and spring-dashpot model.}
    \label{fig:kelvin}
\end{figure}

\subsection{Perturbation Tests}

\begin{figure}[h] 
\centering 

\begin{subfigure}[h]{0.3\textwidth} 
    \includegraphics[width=\textwidth]{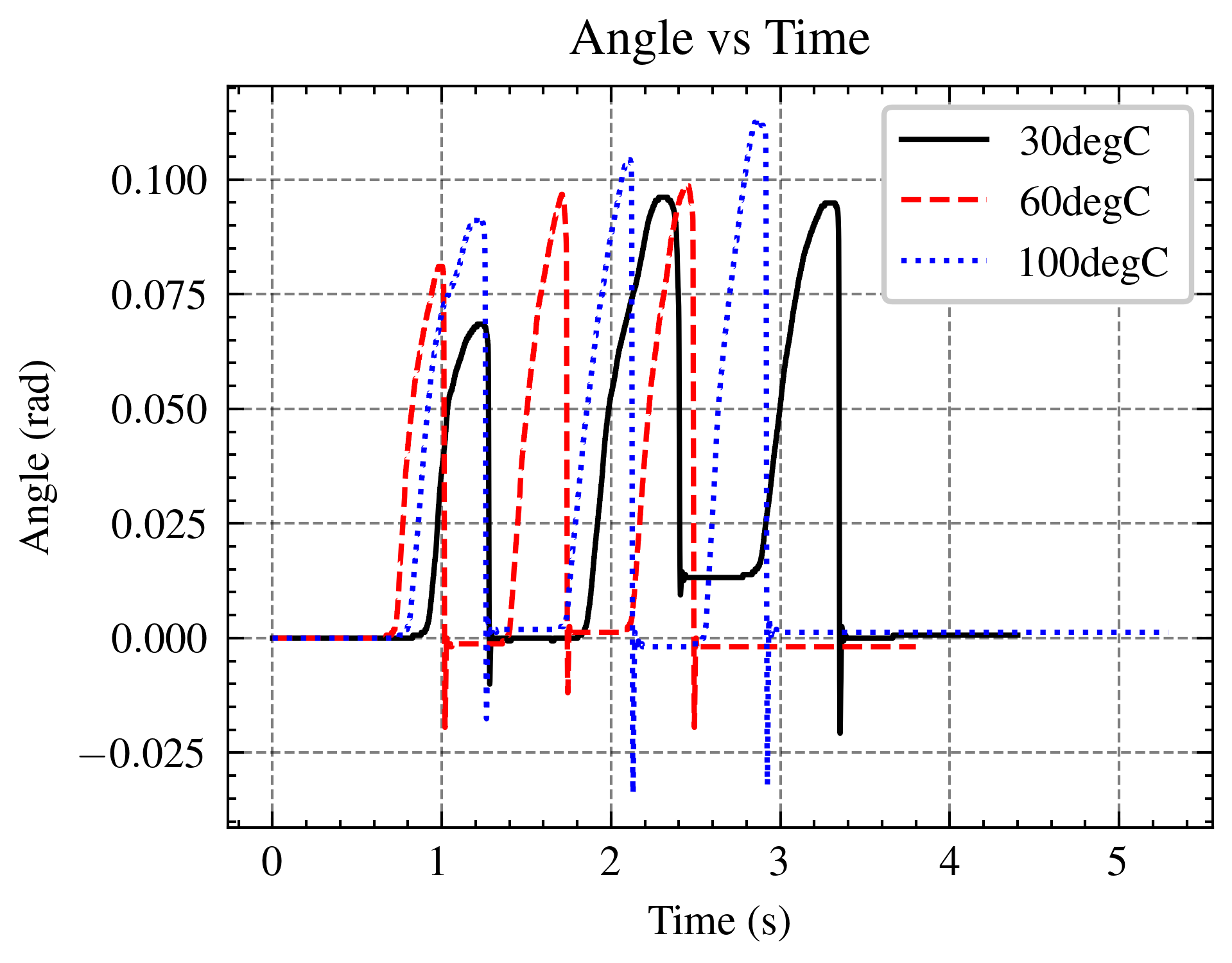} 
    \caption{Output link angle vs time.}
    \label{fig:angletime}
\end{subfigure}
\hfill 

\begin{subfigure}[h]{0.3\textwidth}
    \includegraphics[width=\textwidth]{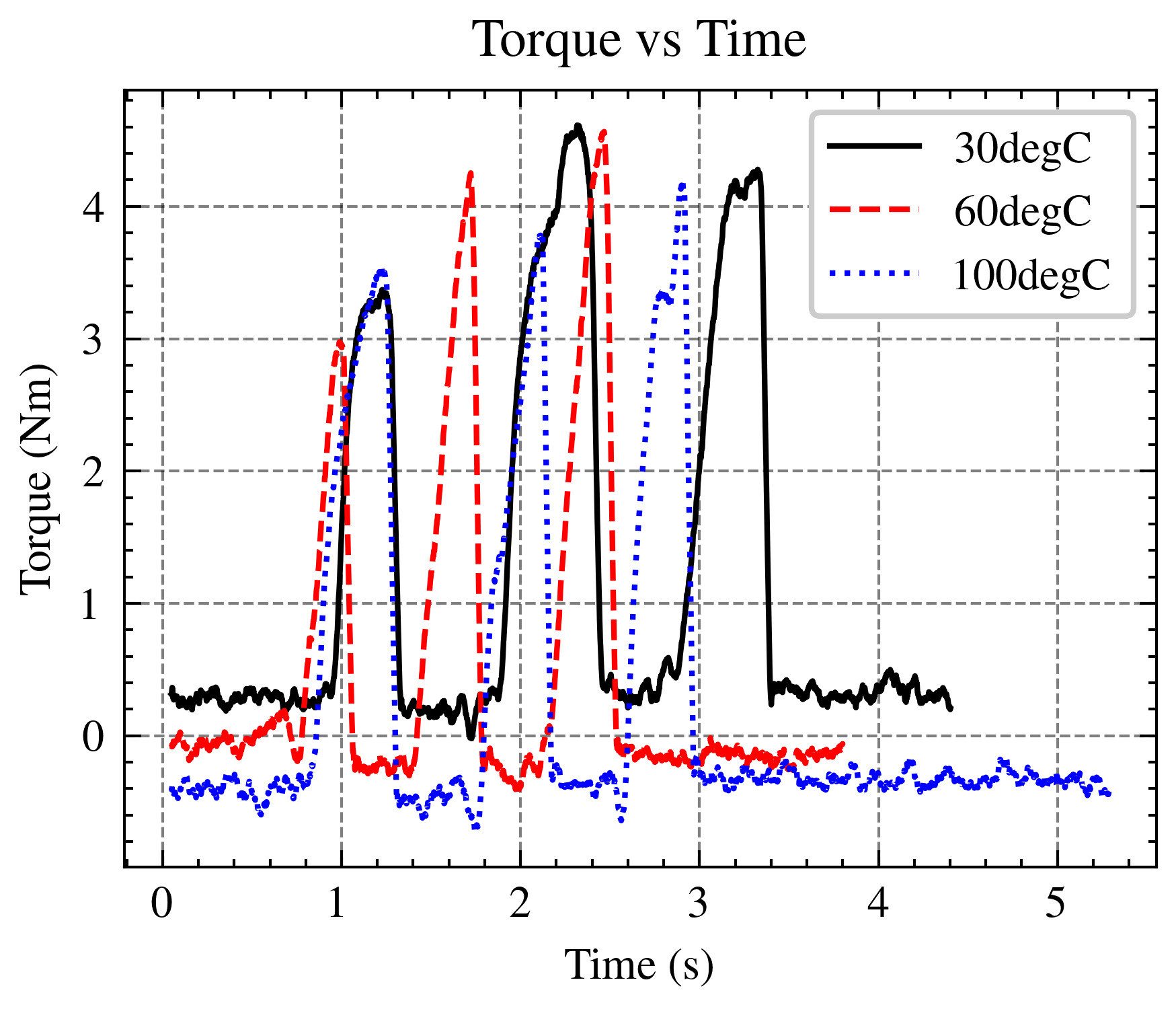}
    \caption{Output link torque vs time.}
    \label{fig:torquetime}
\end{subfigure}
\hfill 

\begin{subfigure}[h]{0.3\textwidth}
    \includegraphics[width=\textwidth]{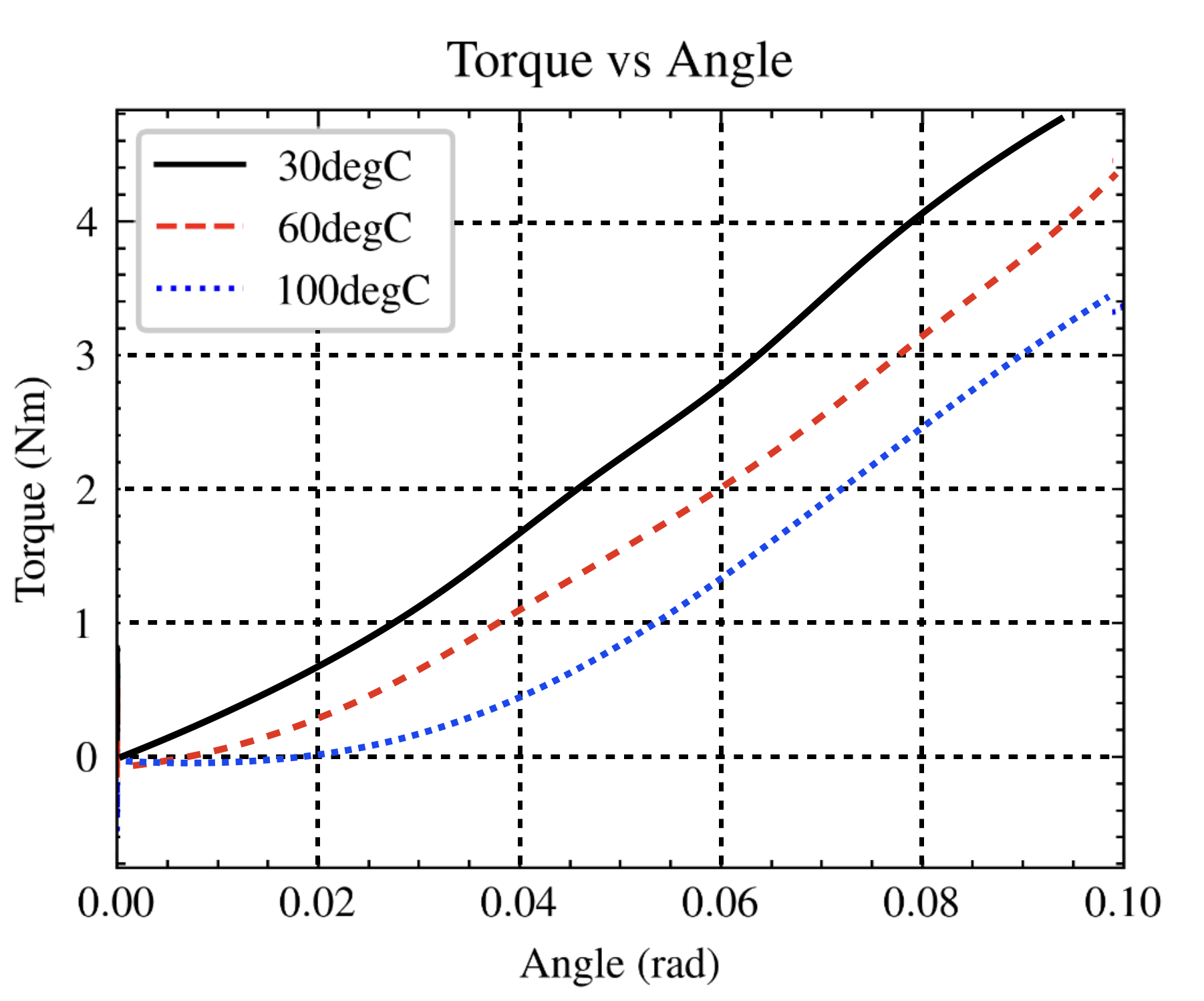}
    \caption{Output link torque vs angle}
    \label{fig:torqueangle}
\end{subfigure}

\caption{Graphs of perturbation testing} 
\label{fig:perturb}
\end{figure}

The VIM was attached to an optical table (Thorlabs; NJ, USA) and the output link was deflected to 0.1 rad and let go while recording torque and angle as it returned to the equilibrium position. The perturbation testing of the VIM was conducted at 30$^\circ$C, 60$^\circ$C, and 100$^\circ$C. From these initial results, we can find the variable stiffness of the shear-mode design in Fig. \ref{fig:torqueangle}. These results corroborate our previous findings from the compression-mode design

\section{Discussion}

The rheology results shows that the dynamic moduli of PCL are frequency-dependent at temperatures higher than 60$^\circ$C. This is noteworthy as the storage and loss moduli are orders of magnitude lower at frequencies lower than 1 Hz than at higher frequencies ($>$100 Hz). This indicates that the variable stiffness behavior of the PCL is more evident in high-bandwidth cases, while the variable damping is dominant at low-bandwidth perturbations.

The transient heating of the PCL was improved from the compression-mode design, by decreasing the total distance the heat has to travel from the Peltier to affect the entire PCL body (shear mode takes 70 s at 1.5 A to reach 100$^\circ$C from room temperature, while compression-mode takes 140 s for the same).

The shear-mode VIA has not been fully realized, as a final design would incorporate a direct-drive motor with an integrated rotary encoder. Additionally, the entire housing and rotor of the VIA will be machined out of aluminum.

\begin{figure}[h] 
\centering 

\begin{subfigure}[h]{0.95\columnwidth} 
    \includegraphics[width=\textwidth]{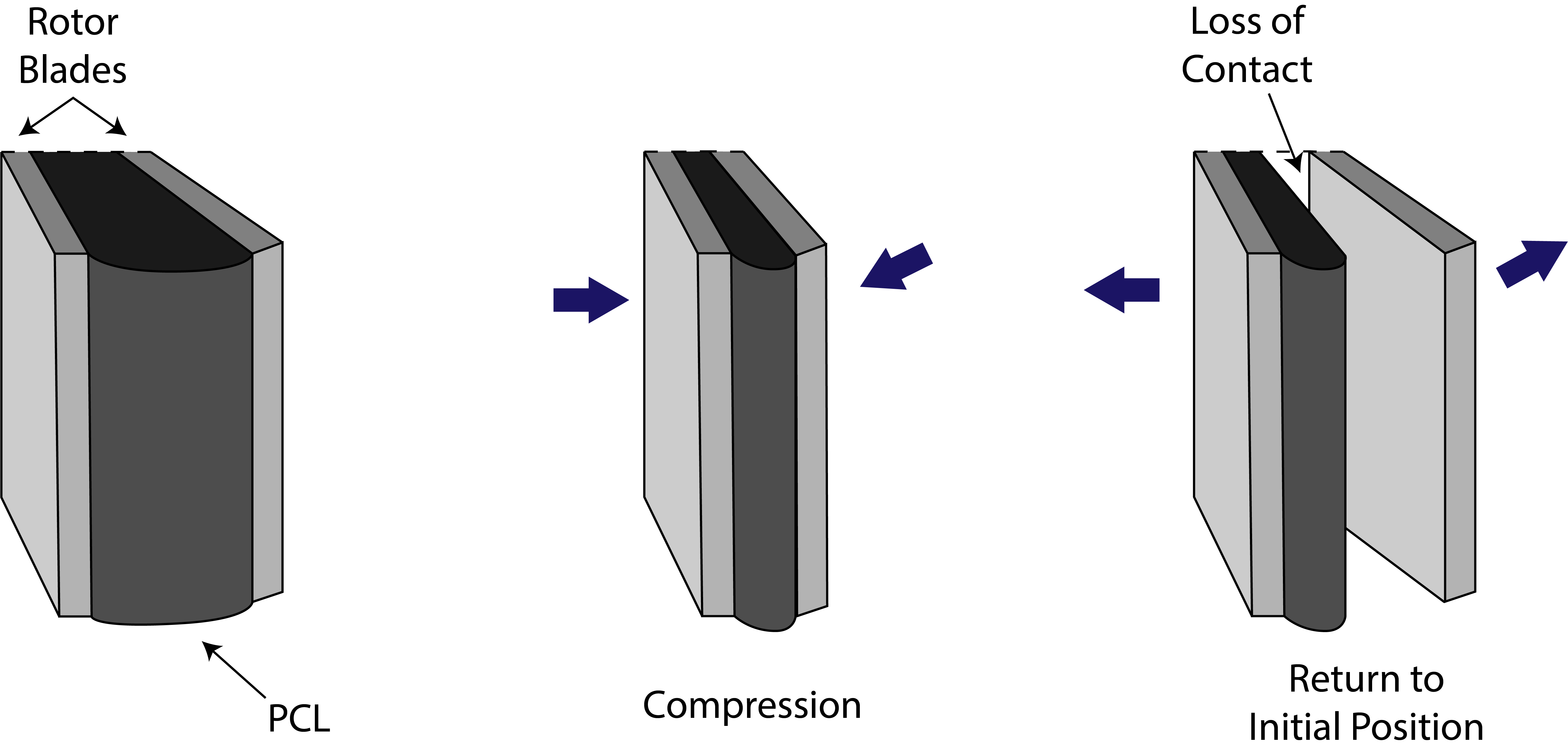} 
    \caption{Compression-mode drawback showing PCL losing contact during perturbations.}
    \label{fig:compmode}
\end{subfigure}
\hfill 

\begin{subfigure}[h]{0.95\columnwidth}
    \includegraphics[width=\textwidth]{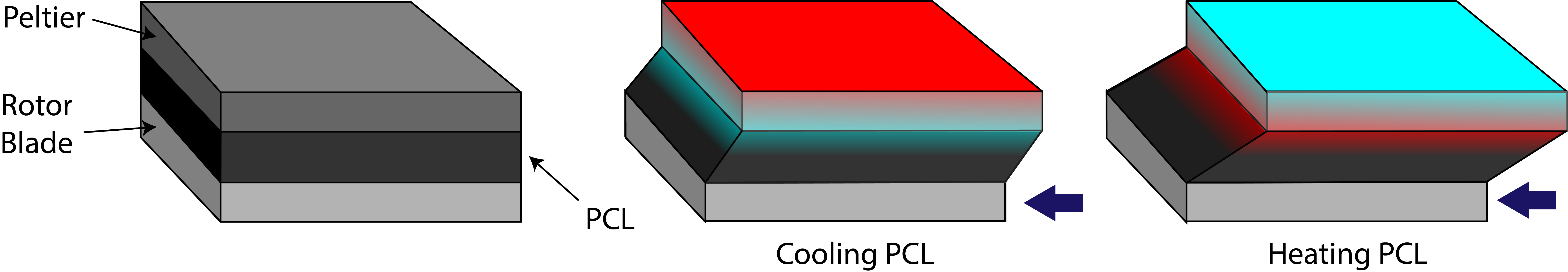}
    \caption{Shear-mode showing full contact of PCL during perturbations, and localized heating from the flexible Peltiers.}
    \label{fig:shearmode}
\end{subfigure}
\hfill 

\caption{Diagrams showing improvement of shear-mode design in eliminating backlash and ensuring more efficient heat transfer.} 
\label{fig:compshear}
\end{figure}

\section{Conclusion}
This study advances the field of variable impedance actuators by introducing an improved thermo-active design that leverages the shear-mode operation of polycaprolactone for enhanced impedance control. By integrating a torque sensor, torsion springs, and strategically placed thin polycaprolactone arcs in close proximity to flexible Peltier elements, we have developed a more responsive and precise actuator system. This novel approach not only maintains the simplicity and compactness of our initial design but also addresses its limitations in stress-relaxation and backlash. Our prototype demonstrates a significant step forward in achieving more efficient and fine-tuned control over both the elastic and viscous properties of the actuator in an off-line setting.

Looking ahead, we aim to further enhance the actuator's performance by integrating a frameless direct drive motor. This will enable comprehensive testing under sinusoidal and step trajectory conditions to closely examine the actuator's impedance responsiveness in real-time applications. Once the motor is integrated and dynamic testing conducted, we will implement into the VIA into a small-form factor \textit{p}HRI application joint.


%

\section*{Acknowledgment}

This work was funded by National Science Foundation NSF under Grant Number 2045177, and by the National Institute of Health (NIH) through grant T32GM136501.

\ifCLASSOPTIONcaptionsoff
  \newpage
\fi

\bibliographystyle{IEEEtran}

\bibliography{biblio}

\end{document}